\documentclass[11pt]{article}

\usepackage[preprint]{acl}

\usepackage{times}
\usepackage{latexsym}
\usepackage{amsfonts}
\usepackage{amsmath}
\usepackage{booktabs}
\usepackage{multirow}
\usepackage{multicol}
\usepackage{enumitem}
\usepackage[capitalize,noabbrev]{cleveref}
\usepackage{xspace}
\usepackage{xcolor}

\usepackage{fvextra}                       % fancyvrb + in-line line breaking
\usepackage[framemethod=TikZ]{mdframed}    % framed, breakable boxes

\mdfdefinestyle{promptstyle}{
  backgroundcolor=gray!4,
  linecolor=black!72,
  linewidth=0.6pt,
  roundcorner=2pt,
  innerleftmargin=6pt, innerrightmargin=6pt,
  innertopmargin=6pt,  innerbottommargin=6pt,
  frametitlebackgroundcolor=black!72,
  frametitlefont=\normalfont\bfseries\color{white},
  frametitlerule=false,
}

\usepackage[T1]{fontenc}
\usepackage[utf8]{inputenc}

\usepackage{microtype}

\usepackage{inconsolata}

\usepackage{graphicx}

\title{\method{}: Continuous Theory-of-Mind Tokens for \\ Multi-Agent Coordination}

\author{First Author \\
  Affiliation / Address line 1 \\
  Affiliation / Address line 2 \\
  Affiliation / Address line 3 \\
  \texttt{email@domain} \\\And
  Second Author \\
  Affiliation / Address line 1 \\
  Affiliation / Address line 2 \\
  Affiliation / Address line 3 \\
  \texttt{email@domain} \\}

\newcommand{\method}[0]{\textsc{ToM2Vec}\xspace}

\author{
    \textbf{Nithin Sivakumaran}\textsuperscript{1} \quad\quad
    \textbf{Ali Payani}\textsuperscript{2} \quad\quad
    \textbf{Hyunji Lee}\textsuperscript{1} \quad\quad \\
    \textbf{Mohit Bansal}\textsuperscript{1} \quad\quad
    \textbf{Elias Stengel-Eskin}\textsuperscript{3} \\
    \textsuperscript{1}UNC Chapel Hill \quad \textsuperscript{2}Cisco \quad \textsuperscript{3}The University of Texas at Austin
}

\begin{document}
\maketitle
\begin{abstract}
Forming an accurate theory-of-mind (ToM) is key to successful multi-agent collaboration and requires reasoning under uncertainty to capture multiple possible partner states.
This is hard to do in the text representation space native to large language model (LLM) agents, which more easily describe a single partner state.
To address this challenge, we introduce \method{}, which learns to form abstract ToMs in a rich latent space via a fixed budget of continuous reasoning tokens. 
During training, supervision for these tokens comes from the partner's actual thoughts, 
which the ToM tokens are trained to abstract and match. 
At test-time, ToM tokens are generated by each agent and act as conditioners for that agent's thoughts and actions. 
Across several environments covering collaborative, competitive, and mixed-motive settings (Overcooked-AI, multi-party negotiation, Decrypto, Hanabi, and Collab Escape/Capture), we evaluate \method{}'s ability to support multi-agent interactions and collaboration, finding consistent improvements over the base model and a variety of competitive trained and prompted baselines.
We show that latent reasoning is critical, with performance dropping when reasoning is expressed in text. 
Moreover, we show that \method{} training generalizes to partner model families not seen at train-time, including a proprietary model (GPT-5.4).
In our analysis, we provide evidence that the latent tokens encode a ToM, rather than task-specific information:  \method{}'s tokens improve the agent's ability to directly answer questions about partner beliefs, desires, and next actions.\footnote{Code: \url{https://github.com/nsivaku/tom2vec}}
\end{abstract}

\section{Introduction}
\label{sec:intro}

Humans have the ability to coordinate with each other remarkably well, often with little or no explicit communication. For instance, chefs can prepare a meal together, teammates can execute a play, and negotiators can converge on a deal, each anticipating what the other will do and acting accordingly. This capacity for joint action rests on each agent's Theory of Mind (ToM): the ability to model another agent's unobserved mental state based on observed behaviors \citep{premack1978does}. 
Large language models (LLMs) are increasingly deployed as agents that must act alongside partner agents (human or AI) in both collaborative and competitive tasks 
\citep{duan2024gtbench, chen2024reconcile, stengellacie, kosinski2024evaluating, xu2024opentom, zhou2025tom, xiao2026playing}. To succeed in these settings, an agent needs strong working ToM \citep{meta2022human, li2023theory, chen2025theory}.  
This ToM allows the agent to infer a partner's goals, beliefs, and likely next actions, enabling it to coordinate with less explicit communication and adapt to a counterpart's strategy on the fly \citep{baker2009action, wu2020too}.

Making these kinds of inferences can be seen as a form of world modeling directed at another agent \citep{raileanu2018modeling, xie2021learning, yu2022model,  chernyavskiy2025applying,  zhangmetamind}.
As in world modeling, it is often important to form an abstract model that can superimpose multiple possibilities, in this case of the partner's intent \cite{xie2021learning, chernyavskiy2025applying}.
A general sketch of the partner's beliefs and intentions is often sufficient, and excessive fidelity can be inefficient and in fact harmful to the task \citep{mu2026adaptive}; for example, when two chefs have to coordinate in a crowded kitchen, one does not need to predict what the other wants for breakfast tomorrow (which might even be a distraction), but does need to predict if the other needs shared resources.

\begin{figure*}[ht]
  \begin{center}
    \centerline{\includegraphics[width=\textwidth]{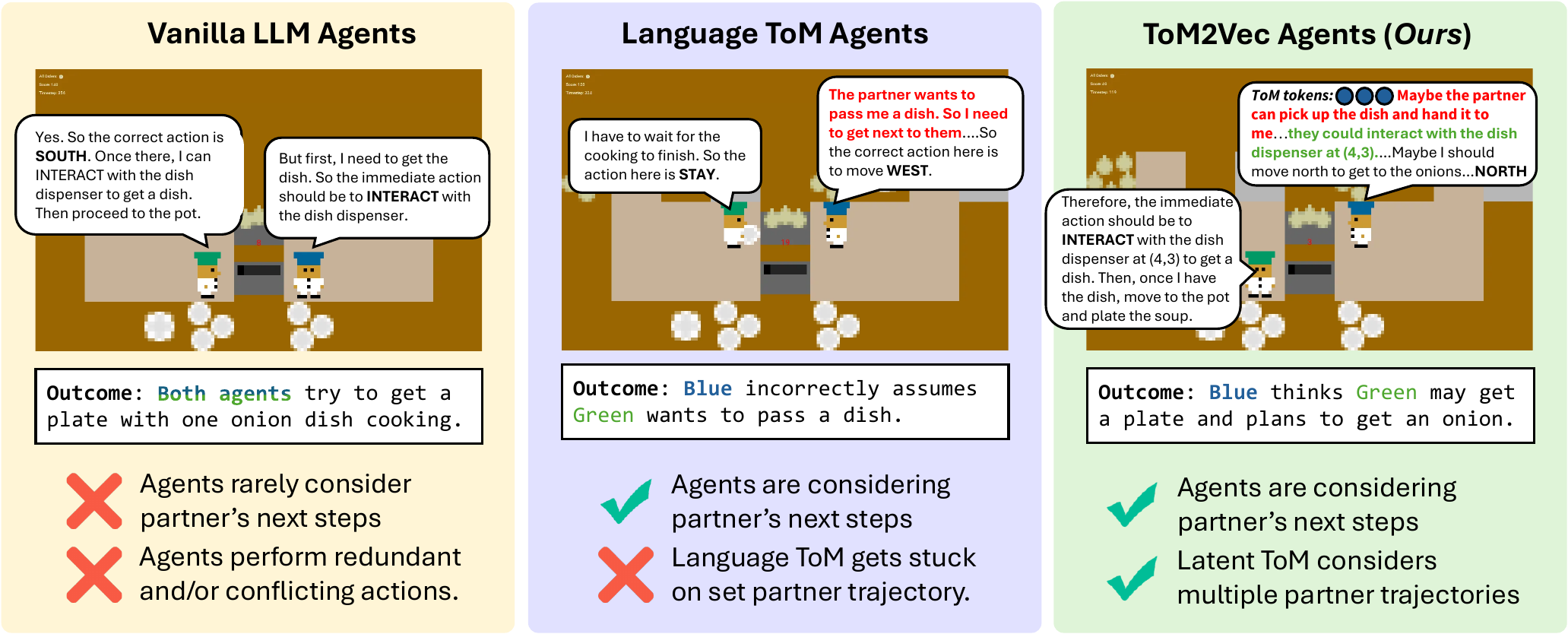}}
    \caption{
      \textit{Left}: Vanilla LLM agents (using Qwen3-8B) each reason only about their own plan. Both independently conclude they should fetch a dish from the dispenser, producing redundant and conflicting behavior. \textit{Middle}: Language ToM Agents are able to more frequently consider their partner's next actions, but they often commit to a single line of reasoning without considering other potential paths. \textit{Right}: Our method \method{} is used to train an agent (blue hat) with stronger ToM capabilities; that agent considers what the untrained partner agent (green hat) is doing before making its own decision.
      } 
      \vspace{-1.5em}
    \label{fig:fig1}
  \end{center}
\end{figure*}

Given this goal of abstraction, a key challenge for ToM agents lies in building an abstract representation
that retains enough information about the partner's potential future paths to support prediction and planning, while lifting unnecessary details.
We argue that natural language is a poor medium for this kind of abstraction, as expressing another agent's intent or state linguistically tends to commit the agent to a single estimate of the partner, even when many distinct hypotheses are possible.
Instead, we argue that modeling another agent may be more akin to forming a latent world model \citep{lecun2022path}, 
where abstractions are formed in the model's latent space rather than its output space (see \cref{fig:fig1}). 
Latent reasoning paradigms offer precisely this kind of latent-space abstraction as a natural alternative to verbalized reasoning, 
where latent thought tokens can superimpose several candidate chains-of-thought rather than fixing to one 
\citep{hao2025training, li2025implicit, chen2025reasoning, shi2026swireasoning}. 
Similarly, we argue that a representation of a partner should be able to superimpose multiple hypotheses about the partner's goals and intentions, and that a ToM agent should be able to predict these representations and consider alternatives before committing to a decision.

\begin{figure*}[ht]
  \begin{center}
    \centerline{\includegraphics[width=\textwidth]{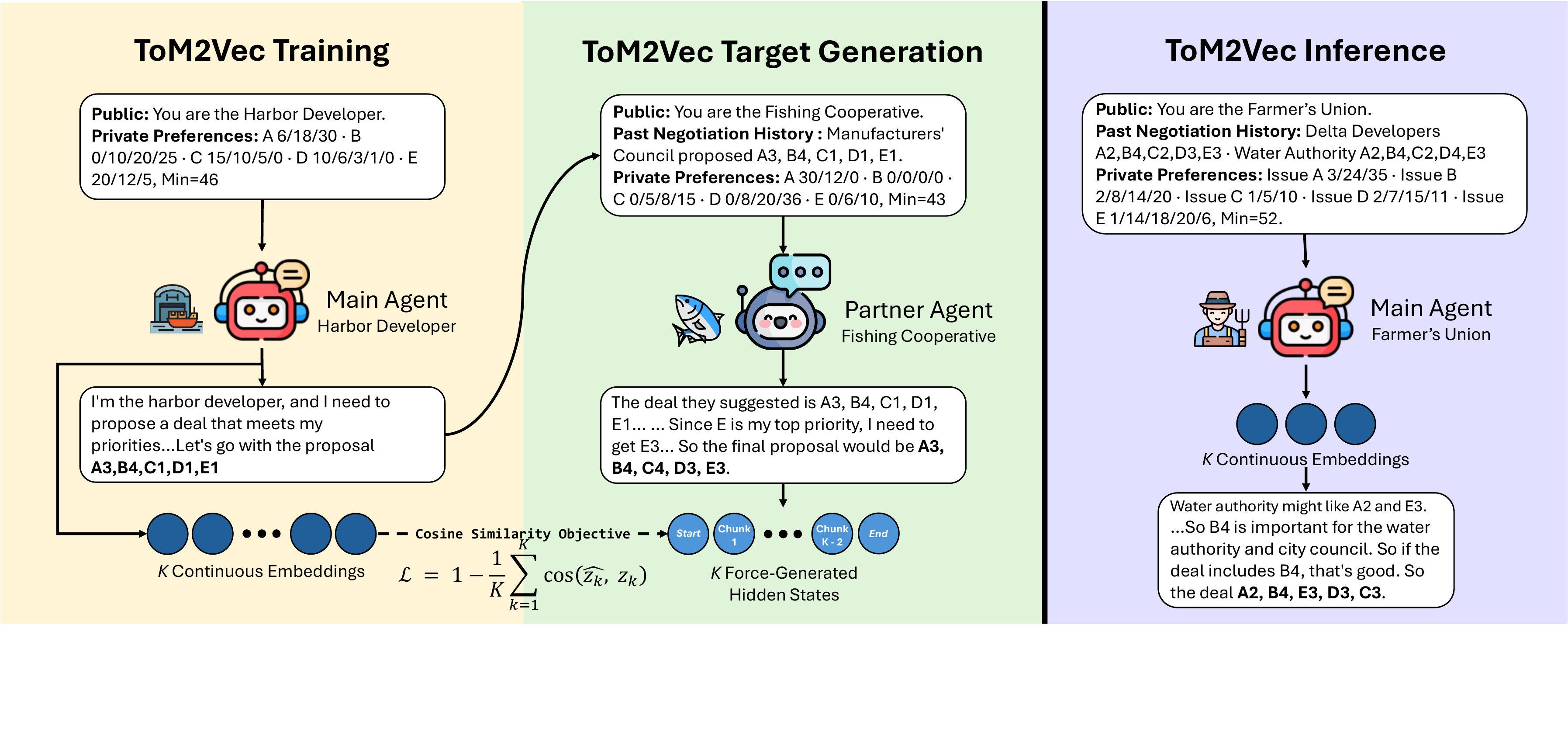}}
    \vspace{-3.25em}
    \caption{
      Overview of \method{} on the multi-party negotiation task.
      \textit{Left (training):} From its local observation and interaction history, the main agent predicts $K$ continuous latent ToM tokens that summarize a partner's likely reasoning trace.
      \textit{Middle (training):} The partner's reasoning is segmented at sentence delimiters, and latent tokens are extracted across the trace.
      These $K$ vectors provide self-supervised targets for position-wise cosine alignment with the main agent's predictions.
      \textit{Right (inference):} The main agent predicts the latent ToM tokens using only its local context and conditions its subsequent reasoning and action on them. The partner's private context and reasoning are used only to construct training targets.
    }
    \vspace{-0.5em}
    \label{fig:method}
  \end{center}
\end{figure*}

To this end, we introduce \method{} to directly learn these latent representations of the partner agent. \method{} trains the agent to predict, from its own point-of-view, a fixed number of continuous latent embeddings that abstract over its partner's thoughts; in other words, it teaches the main agent to perform a mind-reading task about its partner. 
During training, we obtain supervision for these latent embeddings by abstracting a training partner's actual next thoughts into continuous latent tokens to serve as a representation of its reasoning. 
We train the agent's generated latent embeddings to match the partner's hidden states under a cosine similarity objective (see \cref{fig:method}); 
during training, we use the same model for both agent and partner, as their representation spaces must be shared.
At test-time (where partner reasoning is not available) an agent trained with \method{} then generates its own latent thought tokens, intended to capture a ToM about the partner. 
Interestingly, we show that at test-time, our training generalizes to other partner models, including a closed-source proprietary model that is stronger than the training partner.

We evaluate \method{} across cooperative and competitive tasks, including (1) Overcooked \citep{carroll2019utility}, a cooperative cooking environment, (2) multi-party LLM negotiation \citep{abdelnabicooperation}, (3) Decrypto \citep{lupu2025decrypto}, a ToM task where two agents attempt to pass secret messages without being decoded by an interceptor agent, (4) Hanabi \citep{agashe-etal-2025-llm}, a cooperative card game with hidden information, and (5) Collab Escape/Capture \citep{agashe-etal-2025-llm}, two  collaborative coordination tasks. 
Across these domains, \method{} delivers consistent improvements over competitive baselines. Relative to the untrained model (Qwen3-8B), training with \method{} improves the average Overcooked score by 12.0 points and raises negotiation deal success from 62.0\% to 71.0\%. It also outperforms approaches that prompt agents to consider their partners' intentions or train them to explicitly verbalize their partners' reasoning, supporting the value of learning latent partner representations. In Overcooked, these benefits extend to unseen partners from different model families: \method{} improves over the base agent by 17.3 points on average with a GPT-5.4 partner and matches or exceeds the base agent with a Gemma-4 partner, despite training only with a Qwen3-8B partner. 

Our analysis further examines to what extent \method{} learns representations that capture a partner's evolving beliefs. We directly test whether these representations better support inferences about a partner's beliefs, desires, and next actions (i.e., \citet{baker2011bayesian}'s Bayesian ToM modeling) 
within a given environment state. 
We also test whether the latent tokens encode partner-specific reasoning by deliberately corrupting the context used to derive them. Latents derived from the partner's actual context assign lower perplexity to its true reasoning than those derived from an altered layout or an adversarial intent, supporting the interpretation that they capture context-sensitive information about the partner rather than only generic signals useful for coordination. 
We complement this analysis with additional results in the appendix, showing that \method{} helps in Sally-Anne-style false-belief tests, and that the representations learned by \method{} support belief revision given new evidence. 

\section{Method}

\subsection{Problem Setup}

We formulate the multi-agent task as a decentralized partially observable Markov decision process,
\[
\mathcal{D}
  = \big(\mathcal{I},\mathcal{S},\{\mathcal{A}_i\}_{i\in\mathcal{I}},
  T,\{\mathcal{O}_i\}_{i\in\mathcal{I}},O,R,\rho_0,H\big),
\]
  
\noindent where $\mathcal{I}=\{1,\ldots,n\}$ is the agent set, $\mathcal{S}$ is the state space, $\mathcal{A}_i$ and $\mathcal{O}_i$ are agent $i$'s action and observation spaces, $T$ is the transition kernel, $O$ is the joint observation kernel, $R$ is the shared reward, $\rho_0$ is the initial-state distribution, and $H$ is the horizon. At step $t$, the agents choose a joint action $\mathbf{a}_t=(a_t^1,\ldots,a_t^n)$ and the environment transitions according to $T(s_{t+1}\mid s_t,\mathbf{a}_t)$. Turn-based interactions are represented by including a no-op action to inactive agents. Note that assumption of a shared reward applies only to cooperative tasks, not in competitive tasks.

Each agent $i \in \mathcal{I}$ acts from its own action history $h_t^i=(a_0^i,\ldots,a_{t-1}^i).$ We write $c_t^i$ for the model input context constructed from this history, the current state, role-specific instructions, and general task information.

We detail the training method for a main agent $M\in\mathcal{I}$ modeling a partner $P\in\mathcal{I}$. Both agents are initialized from the same base language model $\mathcal{B}$. We denote the trainable main model by $\mathcal{B}_{\theta}$.

\subsection{Modeling Partners via Latent Reasoning}
\label{sec:method-predict}

Under partial observability, the main agent must anticipate its partner's behavior without observing the partner's complete history and reasoning. We train $\mathcal{B}_{\theta}$ to predict a fixed budget of $K$ continuous latent vectors from the main agent's own context:
\[
  \hat{\mathbf{z}}_t
  =\big(\hat z_{t,1},\ldots,\hat z_{t,K}\big)
  =\operatorname{Latent}_{K}(\mathcal{B}_{\theta};c_t^M),
\]
Here, $\operatorname{Latent}_{K}$ denotes $K$ autoregressive steps in the model's hidden space. To perform these steps, we follow recent work on latent reasoning \citep{hao2025training, shen2025codi} and treat the last layer's hidden state at each step as the next step's input embedding, rather than decoding out to the vocabulary.
These latent ToM tokens are intended to summarize the partner's likely reasoning given what the main agent can observe. 

\subsection{Self-Supervised Training via Self-Play}
\label{sec:method-target}
\method{} operates by teaching agents to predict latent vectors of each other's thought traces.
To obtain a signal for this, we leverage self-play: given a main agent (to be trained) and a partner agent (both parameterized by the same model $\mathcal{B}$), we sample a reasoning trace from the partner, compress it into latent tokens, and then supervise the main agent to autoregressively produce latent tokens that match the latent tokens extracted from the partner. 

More formally, as illustrated in \cref{fig:method} (Middle), given a partner agent's context $c_t^P$, the partner generates a full reasoning sequence $r_t^P \sim \mathcal{B}_\theta(c_t^P)$. 
We first segment $r_t^P$ at sentence delimiters and store the hidden state of each token. To obtain $K$ continuous reasoning tokens, we reserve two positions for the start and end tokens and divide the sentences evenly into $K-2$ chunks. 
We take the hidden state for the token at the center of each chunk to get $K-2$ continuous latent tokens. 
We then add the hidden states from the tokens at the start and end of the reasoning trace, resulting in $K$ latent tokens.
These tokens $\mathbf{z}_t = (z_{t,\text{start}}, z_{t,\text{chunk 1}},\ldots,z_{t,\text{chunk K-2}},z_{t,\text{end}}) = (z_{t,\text{1}}, z_{t,\text{2}},\ldots,z_{t,\text{K}})$ will serve as the targets for training the main agent. 

During training, all agents share a model, so we can directly align the main agent's output to the targets extracted from the partner. 
As shown in \cref{fig:method} (Left), we allow the main agent to autoregressively predict $K$ latent tokens, supervising these tokens via a cosine similarity loss:
\vspace{-0.25em}
\begin{equation}
  \mathcal{L}_{\mathrm{ToM}}
  =1-\frac{1}{K}\sum_{k=1}^{K}
  \cos\big(\hat z_{t,k},z_{t,k}\big),
  \label{eq:tom-loss}
\vspace{-0.25em}
\end{equation}
In practice, we use $K=10$. 
Note that the ToM objective in \cref{eq:tom-loss} is \method{}'s only training signal; we do not directly optimize for task rewards. 
Further details are provided in \cref{app:model}.

\subsection{Inference}

At inference time the partner's info ($c^P_t$, $r^P_t)$ is unavailable, so agent $M$ relies solely on its prediction, as shown in \cref{fig:method} (Right). Given its own context, the agent first emits the $K$ latent ToM tokens $\hat{\mathbf{z}}$ to act as an abstract theory of the partner's likely reasoning. 
These tokens then become part of the context for predicting subsequent tokens, allowing the reasoning and actions that follow them to condition on the latent theory:
\[
  (r_t^M,a_t^M)
  \sim\mathcal{B}_{\theta}
  \big(\cdot\mid c_t^M;\hat{\mathbf{z}}_t\big).
\]
Intuitively, a good ToM agent should reconstruct what its partner is about to do well enough to adjust to it before the partner acts.

\section{Experiments and Results}
\label{sec:results}

\subsection{Experimental Setup}
\label{sec:exp_setup}

\paragraph{Environments.}
We evaluate \method{} on five multi-agent environments that span cooperative (Overcooked-AI, Hanabi, Collab Escape/Capture), mixed-motive (negotiation), and competitive (Decrypto) multi-agent settings.
Overcooked-AI \citep{carroll2019utility} is a fully cooperative
cooking game in which two agents share a kitchen and must interleave complementary subtasks (e.g., fetching onions, plating, and delivering) to maximize the total team score. We use three layouts provided in \citet{carroll2019utility} (Cramped Room, Asymmetric Advantages, and Coordination Ring) each run for 30 seeds. 
Multi-party, multi-issue negotiation \citep{abdelnabicooperation} is a mixed-motive task that we evaluate over 100 samples where each agent holds private preferences over a set of issues and must reach an agreement that balances their own self-interest against the broader risk of walking away with no deal. Decrypto \citep{lupu2025decrypto} is a competitive 
ToM communication game, in which a decoder must interpret a teammate's clues relating to provided hint words (which only the decoder and teammate can view) while an interceptor simultaneously tries to recover the same hidden code from those clues; we run Decrypto over 1000 samples. 
We evaluate multiple environments from the LLM Coordination benchmark \citep{agashe-etal-2025-llm}.
Hanabi is a cooperative card game in which players can observe their
teammates' cards but not their own, requiring them to interpret limited
hints to coordinate their actions; we run Hanabi over 100 seeds.
In Collab Capture, two agents
coordinate their movement and actions to capture an adversary.
In Collab Escape, two agents must repair generators to open an exit
while evading an adversary. In all environments, each agent receives only the observation permitted by its role, and cannot see their partners' private contexts. 
Following \citet{agashe-etal-2025-llm}, we evaluate Collab Capture and Collab Escape on 6 and 12 seeds respectively.

\paragraph{Models.} We use Qwen3-8B \citep{yang2025qwen3}, trained with \method{} entirely in self-play with the partner instantiated from the same base model. 
To test the partner generality of the learned ToM, we additionally evaluate against partner agents drawn from other model families at test time, 
including Gemma-4-E4B \citep{gemma} and GPT-5.4 \citep{singh2025openai}, which are not seen during training, and the latter of which is far stronger.

\subsection{Metrics}
\label{sec:metrics}
We report each domain's standard metrics.
For Overcooked, we report mean episode score.
For LLM Negotiation, we report deal-success rate (deal is acceptable for at least 5/6 parties) together with the agent's own score and the average collective score across all parties. 
For Decrypto, we report the decoder win rate (how often the team guesses the codes), and interceptor win rate (how often the interceptor guesses the codes). 
For Hanabi, we report game score, which is the number of correctly played cards. 
For Collab Capture and Collab Escape, we report capture and escape success rates, respectively, together with average turn counts, following \citet{agashe-etal-2025-llm}.
A strong ToM agent should improve across these tasks, which each require some form of partner modeling.

\subsection{Baselines}
We compare \method{} to the following baselines:
\begin{itemize}[noitemsep,topsep=0pt,leftmargin=*]
    \item \textbf{Untrained Base Model}:
    the untrained base agent acting on its own context, with no explicit partner modeling. Note that ``base'' here refers to the untrained nature of this starting point model -- in practice, all models used are post-trained.
    \item \textbf{Prompted ToM}: the agent is explicitly prompted to reason about its partner's observations, goals, and likely next actions before acting.
    \item \textbf{Language ToM Training}: a model trained to verbalize a text summary of the partner's reasoning (i.e., to decode the partner's summarized CoT as discrete text), rather than predicting latent ToM tokens. 
    This isolates the contribution of the latent representation.
    \item \textbf{Self-Consistency} \citep{wang2023selfconsistency}: the agent samples three reasoning chains for its next decision and acts on the majority choice.
    \item \textbf{SimToM} \citep{wilf2024think}: a ToM prompting method in which the agent first recognizes what it knows about the partner, then conditions its own reasoning and action on that inferred perspective.
    \item \textbf{Adaptive ToM} \citep{mu2026adaptive}: a ToM method for Overcooked in which the agent estimates its partner's ToM reasoning order based on prior interactions, to improve its own reasoning.
    \item \textbf{Oracle Partner}: an upper bound in which the agent is given the partner's actual relevant private thoughts (i.e., next action and intent) at inference, obtained from the partner's next turn. 
\end{itemize}

\subsection{Main Results}

\begin{table}[t]
  \centering
  \small
  \resizebox{\columnwidth}{!}{%
  \begin{tabular}{lcccc}
    \toprule
    \textbf{Model} & \textbf{CrR} & \textbf{AA} & \textbf{CoR} & \textbf{Avg.} \\
    \midrule
    Untrained     & $90.0_{\pm{1.86}}$  & $91.3_{\pm{3.28}}$  & $102.7_{\pm{1.26}}$ & 94.7  \\
    \midrule
    Prompted ToM & $86.7_{\pm{1.75}}$  & $96.0_{\pm{3.64}}$  & $106.7_{\pm{1.75}}$ & 96.4  \\
    Language ToM & $92.0_{\pm{2.81}}$  & $98.7_{\pm{0.93}}$  & $102.0_{\pm{1.11}}$ & 97.6  \\
    Self-Consistency & $91.3_{\pm{3.28}}$ & $97.3_{\pm{1.26}}$ & $107.3_{\pm{1.79}}$ & 98.7 \\
    SimToM           & $90.7_{\pm{1.85}}$ & $94.0_{\pm{1.70}}$ & $105.3_{\pm{1.64}}$ & 96.7 \\
    Adaptive ToM & $92.7_{\pm{1.79}}$ & $102.7_{\pm{0.93}}$ & $101.3_{\pm{2.34}}$ & 98.9 \\
    \textcolor{gray}{Oracle Partner$^\dagger$} & \textcolor{gray}{$110_{\pm{2.29}}$} & \textcolor{gray}{$118_{\pm{1.38}}$} & \textcolor{gray}{$125_{\pm{1.99}}$} & \textcolor{gray}{117.7} \\
    \midrule
    \method{} & $\mathbf{98.0}_{\pm{2.22}}$ & $\mathbf{107.7}_{\pm{2.08}}$ & $\mathbf{114.3}_{\pm{2.66}}$ & $\mathbf{106.7}$ \\
    \bottomrule
  \end{tabular}
  }
  \caption{Overcooked mean episode score and standard error on three layouts and their average score across 30 seeds. All settings use Qwen3-8B. We bold the best mean episode score for each layout. CrR, AA, and CoR stand for the Cramped Room, Asymmetric Advantages, and Coordination Ring layouts respectively. $^\dagger$Oracle has privileged access to the partner's context and serves as an upper bound.
  }
  \label{tab:overcooked_main}
\end{table}

\begin{table}[t]
  \centering
  \resizebox{\linewidth}{!}{
  \begin{tabular}{lccc}
    \toprule
    \textbf{Method} & \textbf{Deal \%} & \textbf{Own Score} & \textbf{Collect. Score} \\
    \midrule
    Untrained         & 62.0 & 59.2 & 54.7 \\
    \midrule
    Prompted ToM  & 68.0 & 61.5 & 56.8 \\
    Language ToM  & 64.0 & 60.8 & \textbf{59.2} \\
    Self-Consistency  & 63.0 & 58.4 & 58.0 \\
    SimToM  & 65.0 & 62.1 & 58.4 \\
    \midrule 
    \method{}        & \textbf{71.0} & \textbf{65.6} & \textbf{59.2} \\
    \bottomrule
  \end{tabular}
  }
  \caption{Evaluation results for multi-party, multi-issue negotiation \citep{abdelnabicooperation} across 100  negotiations. Deal \% is the fraction of games ending in an agreement (at least 5/6 agents must agree). Own and Collective Score are the agent's own average final score and the average per-party utility from the final deal.
  }
  \label{tab:negotiation}
\end{table}

\begin{table}[t]
\centering
\resizebox{\linewidth}{!}{
\begin{tabular}{lcc}
\toprule
\textbf{Model} & \textbf{Decoder Win \%} & \textbf{Interceptor Win \%} \\
\midrule
Untrained        & 10.4\% & 87.4\% \\
\midrule
Prompted ToM & 12.8\% & 88.2\% \\
Language ToM & 13.6\% & 93.0\% \\
Self-Consistency & 12.0\% & 90.5\% \\
SimToM & 11.5\% & 92.6\% \\
\midrule
\method{}    & \textbf{14.8\%} & \textbf{93.5\%} \\
\bottomrule
\end{tabular}
}
\caption{Decrypto results: win rates for each method as the Decoder and Interceptor for 1000 samples each. In all settings, the other agents are Qwen3-8B.}
\label{tab:decrypto}
\end{table}

\begin{table}[t]
  \centering
  \small
  \begin{tabular}{lc}
    \toprule
    \textbf{Model} & \textbf{Score} \\
    \midrule
    Untrained        & 8.9 \\
    \midrule
    Prompted ToM     & 9.6 \\
    Language ToM     & 10.6 \\
    Self-Consistency & 9.5 \\
    SimToM           & 10.3 \\
    \midrule
    \method{}        & \textbf{11.3} \\
    \bottomrule
  \end{tabular}
  \caption{Hanabi scores (out of 25). Higher is better, highest score bolded.}
  \label{tab:hanabi}
\end{table}

\begin{table*}[t]
  \centering
  \small
  \begin{tabular}{lcccc}
    \toprule
    & \multicolumn{2}{c}{\textbf{Collab Capture}}
    & \multicolumn{2}{c}{\textbf{Collab Escape}} \\
    \cmidrule(lr){2-3}\cmidrule(lr){4-5}
    \textbf{Model}
    & \textbf{Capture Rate} & \textbf{Avg. Turns}
    & \textbf{Escape Rate} & \textbf{Avg. Turns} \\
    \midrule
    Untrained
    & 0.67 & 5.00 & 0.75 & 8.00 \\
    \midrule
    Prompted ToM
    & \textbf{1.00} & 4.76 & 0.75 & 8.74 \\
    Language ToM
    & 0.67 & 5.52 & 0.67 & 8.24 \\
    Self-Consistency
    & 0.83 & 5.21 & 0.75 & 8.00 \\
    SimToM
    & 0.83 & 5.00 & \textbf{0.92} & \textbf{7.80} \\
    \midrule
    \method{}
    & \textbf{1.00} & \textbf{4.32} & \textbf{0.92} & 8.00 \\
    \bottomrule
  \end{tabular}
  \vspace{-0.25em}
  \caption{Results on Collab Capture and Collab Escape.
  Success rates are reported as fractions. We bold the highest
  success rate and lowest average turn count in each game,
  including ties; fewer turns is better.}
  \vspace{-0.25em}
  \label{tab:collab_games}
\end{table*}

\paragraph{\method{} consistently improves performance across domains.}
We evaluate \method{} on five multi-agent domains spanning  cooperative, mixed-motive, and competitive tasks. Across all tested domains, \method{} improves over the Qwen3-8B base model. 

\noindent\emph{Overcooked.}
On Overcooked \method{} raises mean episode score by 12.0 points 
on average over the untrained model across three layouts (see \cref{tab:overcooked_main}). 
It outperforms all baselines, including training-based baselines (Language ToM) as well as prior work (SimToM, Adaptive ToM) by a substantial margin, with consistent gains across all three configurations. We find that these gains are statistically significant compared over almost all the baselines (see \cref{app:p-values} for tests). 
\vspace{0.2em}

\noindent\emph{Negotiation.}
In the negotiation task, \method{} raises the deal-success rate from
62.0\% (untrained) to 71.0\% while improving both its own and the collective scores for all other agents (see \cref{tab:negotiation}). 
\method{}'s deal success rate and own score are significantly higher than all baselines. 
\vspace{0.2em}

\noindent\emph{Decrypto.}
On Decrypto, \method{} improves win rate in both the cooperative decoder role and the
adversarial interceptor role by 4.4 and 6.1 percentage points respectively~(see \cref{tab:decrypto}), outperforming  
both prompting for ToM and language ToM training (i.e., explicit verbalized ToM prediction). 
\vspace{0.2em}

\noindent\emph{Hanabi.} 
On Hanabi, \method{} achieves a score of 11.3 out of a maximum of 25, improving over the untrained baseline by 2.4 points and the strongest baseline,
Language ToM, by 0.7 points (\cref{tab:hanabi}). Since players must interpret their teammates' hints without observing their own cards, these results suggest that latent ToM can support coordination when agents have access to different information.
\vspace{0.2em}

\noindent\emph{Collab Capture and Escape.}
On Collab Capture, \method{} increases the capture rate from 0.67 to 1.00, a gain of 33 percentage points over the untrained baseline
(\cref{tab:collab_games}). While Prompted ToM also achieves a
capture rate of 1.00, \method{} reduces the average turn count from
4.76 to 4.32, the lowest among the evaluated methods.
On Collab Escape, \method{} increases the escape rate from 0.76 to
0.92, tying the highest success rate while keeping the turn count consistent. 

Overall, the consistency of these results indicates that \method{} learns a partner model that is useful for multi-agent tasks, and that the method generalizes across a wide variety of domains and task configurations. 

\paragraph{Partner information is highly valuable, but
prompting alone fails to capture it.}
Granting the agent oracle access to its partner's private context raises mean score by 23.0 points over the base model, indicating that accurate partner information provides a strong benefit for coordination in these tasks (see \cref{tab:overcooked_main}). 
Prompting a model to reason explicitly about its partner, however, results in more marginal gains (average gain of 1.7 points) 
and even decreases performance on Cramped Room. 
Even more advanced prompting methods like SimToM \citep{wilf2024think} and Adaptive ToM \citep{mu2026adaptive} provide only marginal gains, underscoring the benefits of training.

\paragraph{Latent ToM reasoning outperforms trained natural-language ToM.}
Replacing the latent ToM tokens with an explicit natural-language summary of the partner's reasoning (Language ToM training) recovers only a small part of \method{}'s gains. For Overcooked, Language ToM training averages 97.6 and trails \method{} on all three layouts by 6.0, 9.0, and 12.3 points (see \cref{tab:overcooked_main}). 
In some instances, latent ToM reasoning succeeds where verbalizing the partner's reasoning through text fails. Specifically, on Coordination Ring Language ToM training falls 0.7 points below the base model, whereas \method{} improves it by 11.6. 
Similar trends can be seen across all other domains; in Collab Escape (\cref{tab:collab_games}) Language ToM even leads to a 10 percentage point (pp) drop in performance, whereas \method{} improves by 16pp, highlighting the importance of latent thoughts.

\section{Discussion and Analysis}
\label{sec:analysis}

\begin{table}[t]
  \centering
  \small
  \begin{tabular}{llccc}
    \toprule
    \textbf{Model} & \textbf{Partner} & \textbf{CrR} & \textbf{AA} & \textbf{CoR} \\
    \midrule
    \multirow{2}{*}{Qwen3-8B} & Qwen3-8B  & 89  & 93  & 105 \\
                              & \method{} & 98  & 108 & 113 \\
    \midrule
    \multirow{2}{*}{GPT-5.4}  & Qwen3-8B  & 135 & 193 & 174 \\
                              & \method{} & 152 & 214 & 188 \\
    \midrule
    \multirow{2}{*}{Gemma-4}  & Qwen3-8B  & 65  & 90  & 87  \\
                              & \method{} & 68  & 105 & 94  \\
    \bottomrule
  \end{tabular}
  \caption{Overcooked results for partner agents of different architectures across 20 seeds. Each agent is tested with both Qwen3-8B and \method{}. Neither GPT-5.4 nor Gemma-4 were seen during training. 
  }
  \label{tab:partner_agnostic}
\end{table}

\paragraph{The gains are partner-agnostic and transfer across model families.}
Although \method{} is trained purely in self-play against a Qwen3-8B partner, 
the observed gains on Overcooked persist when \method{} is paired with unseen test-time partners (\cref{tab:partner_agnostic}). 
With a strong GPT-5.4 partner, \method{} improves over the base agent by 17.3 points on average, with positive gains on every layout. With a Gemma-4 partner the improvement is smaller but still persists.
This indicates that \method{} learns a generalized model of partner reasoning rather than overfitting to its training-time partner.
This also highlights the generalizability of \method{} to frontier model settings: we can train an agent against open-weight models and deploy it in teams with frontier models.

\begin{table}[t]
  \centering
  \small
  \resizebox{\linewidth}{!}{

  \begin{tabular}{lccc}
    \toprule
    \textbf{Model} & \textbf{Beliefs (\%)} & \textbf{Desires (\%)} & \textbf{Actions (\%)} \\
    \midrule
    \textcolor{gray}{Random} & 16.7 & 16.7 & 16.7 \\
    Untrained                & 41.7 & 53.0 & 55.3 \\
    \midrule
    Prompted ToM             & 40.0 & 55.0 & 56.3 \\
    Language ToM             & 46.7 & 59.3 & 61.3 \\
    Self-Consistency         & 44.0 & 54.7 & 57.7 \\
    SimToM                   & 43.7 & 54.7 & 58.0 \\
    Adaptive ToM             & 45.7 & 60.3 & 60.0 \\
    \midrule
    \method{}                & \textbf{52.3} & \textbf{62.3} & \textbf{64.0} \\
    \bottomrule
  \end{tabular}
  }
  \caption{Accuracy on Bayesian ToM probes in Overcooked. \emph{Beliefs}: what the partner believes the agent is trying to do (a second-order inference). \emph{Desires}: which subtask the partner is currently pursuing. \emph{Actions}: which primitive action the partner will take next. 
  }
  \label{tab:btom}
\end{table}

\paragraph{\method{} infers partner beliefs, desires, and actions more accurately.}
Coordination scores alone do not reveal whether the latent tokens encode the components of a theory-of-mind: it could be that some improved performance comes from learning to predict information correlated with partner actions, without formulating a proper theory of the partner. 
Thus, following the Bayesian Theory of Mind decomposition of an agent into beliefs, desires, and actions \citep{baker2011bayesian}, we directly check each agent's model of its partner with three questions at sampled Overcooked states (\cref{tab:btom}). 
\textit{Belief} questions ask the agent what it thinks its partner thinks the agent (i.e., itself) is trying to do. \textit{Desire} questions ask the partner what it currently wants to do. 
\textit{Action} questions ask which action the partner will take next. 
Beliefs, desires, and actions each have six different options (based on the possible actions/steps within the Overcooked environment)
to choose from, making this a multiple-choice QA task, and test 300 samples per questions (see \cref{app:prompt} for prompts). 
We obtain ground truth labels by 1) continuing the partner's generation and prompting it for its beliefs and desires and 2) directly taking the partner's next action from its output. \method{} achieves the highest accuracy on all three questions (\cref{tab:btom}), improving over the next-best baseline by 5.6, 2.0, and 2.7 points. 
The beliefs category is  unsurprisingly the hardest question type for every method as it is a second-order ToM understanding capability (i.e., the agent must infer its partner's theory about it). 

\begin{table}[t]
  \centering
  \small
  \begin{tabular}{lcc}
    \toprule
    \textbf{Thought Type} & \textbf{PPL ($\downarrow$)} & \textbf{Lowest PPL Rate} \\
    \midrule
    Normal             & 1.412 &  98.6\% \\
    Altered Layout     & 1.702 &  0.6\% \\
    Adversarial Intent & 1.613 & 0.8\% \\
    \bottomrule
  \end{tabular}
  \caption{Perplexity of generated thoughts when trained by \method{} across different types of thoughts. We test on the normal thoughts, altered layout (swapped real layout for alternate layout), and adversarial intent (partner agent is attempting to sabotage).}
  \label{tab:distractor}
\end{table}

\paragraph{Latent ToM tokens encode valid partner-specific reasoning.}
We compare the perplexity the model assigns to the partner's true reasoning when the ToM latents are derived from the partner's actual context against two deliberately corrupted contexts: Altered Layout, in which the current layout is replaced with a game state from another layout, and Adversarial Intent, in which the partner aims to hinder progress. The key assumption is that if the latents are capturing the partner's reasoning, the generated thoughts from the true context should yield the lowest perplexity.
We find that the Normal latents reach a perplexity of 1.412, whereas both distractors are higher, at 1.702 and 1.613 for Altered Layout and Adversarial Intent respectively (see \cref{tab:distractor}). Furthermore, the lowest perplexity is assigned to the correct CoT in 98.6\% of cases with the true partner reasoning.
Corrupting the partner's context degrades how well the latents predict its reasoning, indicating that the latents are sensitive to valid partner-specific information.

\section{Related Work}

\paragraph{Theory of Mind for LLM Agents.}
Theory of Mind is the capacity to infer the mental states of others, and is a cornerstone of human social cognition \citep{premack1978does}. 
A long line of work explores ToM in non-LLM agents \citep{gmytrasiewicz2005framework, pynadath2005psychsim, han2018learning, sequeira2025tomcat, he2025latent}, which we do not discuss in detail, as our scope is LLM agents.
In that context, a large body of work tested the ToM capabilities of varying models \citep{le-etal-2019-revisiting, sap-etal-2022-neural, ullman2023large,   kosinski2024evaluating, xu2024opentom, kim2023fantom}, showcasing where LLMs succeed and fail in different aspects of ToM.
Many past efforts have aimed to improve  LLM ToM through inference-time prompting and decomposition, including perspective-taking prompts \citep{moghaddam2023boosting, wilf2024think} and structured decomposition of the inference into sub-problems \citep{jung2024perceptions, hou2024timetom, sarangi-etal-2025-decompose}.
A related line of work augments models with explicit symbolic belief-tracking structures \citep{sclar2023minding, xu2025enigmatom, tang2024tom}.
Outside of inference-time approaches, other previous approaches train LLMs to improve ToM \citep{sclar2025explore, hwang2025infusing}.
In contrast to these efforts, \method{} learns a latent partner representation supervised directly during self-play to facilitate broader understanding of the partner.

\paragraph{Latent Reasoning Tokens.}
A growing body of work has explored 
latent reasoning.
A paradigmatic example of this is Coconut \citep{hao2025training}, which feeds the model's last hidden state back as the next input embedding, allowing a single step to support several alternative continuations.
Subsequent methods have enabled latent thinking by adjusting how latent thoughts are trained \citep{shen2025codi, wu2025parallel, su2025token}.
Other attempts keep the model frozen and generate soft tokens that are projected into the LLM's representation space \citep{xu2025softcot, xu2025softcot++} or use probability-weighted embeddings \citep{zhang2026soft}. 
Additionally,  \citet{gozeten2025continuous} show how continuous CoTs enable parallel exploration better than discrete CoTs.
Our work builds on prior work training latent reasoning models, especially findings that latent reasoning tokens can encode multiple paths. 
\method{} is, to our knowledge, the first to explore this paradigm for modeling other agents, which poses distinct challenges (e.g., partial observability of other agents). 

\section{Conclusion}
We introduced \method{}, a ToM method that trains an agent via self-play to model its partner via continuous thought tokens. 
Across five settings, both cooperative and competitive, \method{} improves over all baselines and generalizes to partners not seen during training. 
Analysis indicates that \method{} is able to make more accurate inferences about the partner's state, indicating stronger ToM understanding.
Taken together, the results suggest that modeling another agent in latent space is a useful alternative to doing so in text.

\section*{Limitations}

\method{} requires access to the partner's CoT during training. 
However, we show that this CoT does not need to come from the eventual test-time partner: training against one partner transfers to others.
Nevertheless, it is still a requirement of the training procedure, and our supervision relies on both agents sharing a representation space, which is why training uses the same base model for agent and partner.
\method{} also requires a training phase, unlike prompting-based approaches that work at inference time. 
We find this cost is offset by performance, since the method outperforms both prompting and a training-matched verbalized baseline.
Finally, we note that our experiments use English-only tasks, a limitation shared with much prior work on LLM theory of mind and constrained in part by available models and datasets. 

\section*{Acknowledgments}
This work was supported by NSF-CAREER Award 1846185, NSF AI Engage Institute DRL-2112635, National Institutes of Health (NIH) under other transactions 1OT2OD038045- 01, Cisco Faculty Award, and Coefficient Giving. The views and conclusions contained in this document are those of the authors and should not be interpreted as representing official policies, either expressed or implied, of the NIH or other sponsors.

\bibliography{custom}

\appendix
\crefalias{section}{appendix}

\section{Model Details}
\label{app:model}

We initialize from the default generation configuration from HuggingFace for Qwen3-8B (Apache-2.0 license). For Qwen3-8B, we set \texttt{temperature} of $1.0$, \texttt{top\_p} to $0.95$, \texttt{top\_k} to $20$, \texttt{repetition\_penalty} to $1.0$, and \texttt{presence\_penalty} to $1.5$. For Gemma-4-E4B, following the recommended settings, we set \texttt{temperature} to $1.0$, \texttt{top\_p} to $0.95$, and \texttt{top\_k} to $64$. We use vLLM \citep{kwon2023efficient} as the inference library to run both of these models, with \texttt{GPU memory utilization} set to $0.90$. Finally, when testing GPT-5.4, we set \texttt{reasoning effort} to high and \texttt{verbosity} to medium.

We train \method{} with the AdamW optimizer, using a learning rate of $10^{-6}$. We use a batch size of $4$ with a single gradient-accumulation step.

\section{Dataset Details}
\label{app:dataset}

\begin{itemize}
    \item \textbf{Overcooked-AI} (MIT License): For each baseline, we run for 400 steps on the Cramped Room, Asymmetric Advantage, and Coordination Ring layouts for 30 different seeds. For training, we conduct self-play with a random agent to collect 4000 input game states to use during training across different timesteps.
    \item \textbf{Negotiation} (MIT License): We generate 500 new negotiation settings with GPT-5.4 following the new task generation prompt in \citep{abdelnabicooperation}. Following the original paper, we allow discussion to  proceed for 24 turns and take each turn except the last as a training sample. We conduct an 80/20 split on these samples to construct our training and validation data.
    \item \textbf{Decrypto} (Creative Commons Public License): The original paper provides a list of 640 words to create a set of three keywords. For training, we create 1000 randomly sampled keyword sets and simultaneously use decoder and interceptor roles for training. For evaluation, we create another 1000 randomly sampled keyword sets.
    \item \textbf{Hanabi} (MIT License): We train \method{} over 4000 randomly generated Hanabi game states for input and do evaluation across 100 randomly initialized games of Hanabi.
    \item \textbf{Collab Escape/Capture} (MIT License): We conduct self-play with random agents to to get input environment states for training for both Escape and Capture. We train \method{} over 4000 input states for each setting and, following the evaluation scheme in the original paper, evaluate Collab Escape and Collab Capture on 12 and 6 seeds respectively.
\end{itemize}

\section{Additional Results}

\begin{table}[t]
  \centering
  \small
  \begin{tabular}{lcc}
    \toprule
    \textbf{Model} & \textbf{Accuracy} & \textbf{State Understanding} \\
    \midrule
    Untrained            & 0.74 & 0.88\\
    Prompted ToM         & 0.82 & 0.86\\
    Language ToM         & 0.86 & 0.90\\
    Self-Consistency & \textbf{0.88} & \textbf{0.94}\\
    SimToM               & 0.84 & 0.88\\
    Adaptive ToM               & 0.80 & 0.92\\
    \midrule
    \method{}            & \textbf{0.88} & 0.88\\
    \bottomrule
  \end{tabular}
  \caption{Results on the Sally-Anne belief test for different baselines. We present the accuracy as the proportion of output responses that the overseer agents correctly recognized where the first player would look. We also show the state understanding to illustrate how well each method keeps track of each object.}
  \label{tab:sally_anne}
\end{table}

\paragraph{\method{} obtains the strongest performance on ToM belief probing.}
A more classic test of ToM is through reasoning on false agent beliefs. 
More concretely, whether an agent can recognize that another agent holds a belief that differs from the true environment state. We evaluate this with a Sally-Anne style belief test \citep{baron1985does}. To test this, we create a new Overcooked layout with custom ``cupboard'' tiles to test an agent's ToM capabilities. 

A cupboard can be interacted with to store/retrieve an item, and the contents of a cupboard are not visible at all times (i.e., the content must be retrieved to reveal the stored item). We place eight cupboards along the edge of the layout and fill the remaining edge tiles with walls. To conduct this test, we have an overseer agent observe two players within the environment act following a set pattern. Player One takes three unique ingredients and stores each in three different cupboards. Player One exits and can no longer observe the scene. Player Two is given the item locations and chooses three random cupboards to move the ingredients to. Player Two exits, and player one enters. Now, given the game's state history, we ask the overseer agent where it thinks Player One will look for each ingredient. We also independently ask the overseer agent where it believes the ingredients are after seeing them rearranged by player two to measure state understanding.

We conduct this test on 50 different seeds and find that \method{} scores 0.88 (\cref{tab:sally_anne}), a 0.14 improvement over the untrained agent, tying Self-Consistency for the best result and outperforming all other baselines. We also note that state understanding stays relatively consistent among the different methods, differing by 0.08 at most. Since all the methods retain similar levels of state understanding, the gains in partner state understanding are due to enhanced ToM capabilities rather than better overall state understanding.

\begin{figure}[t]
  \begin{center}
    \centerline{\includegraphics[width=\linewidth]{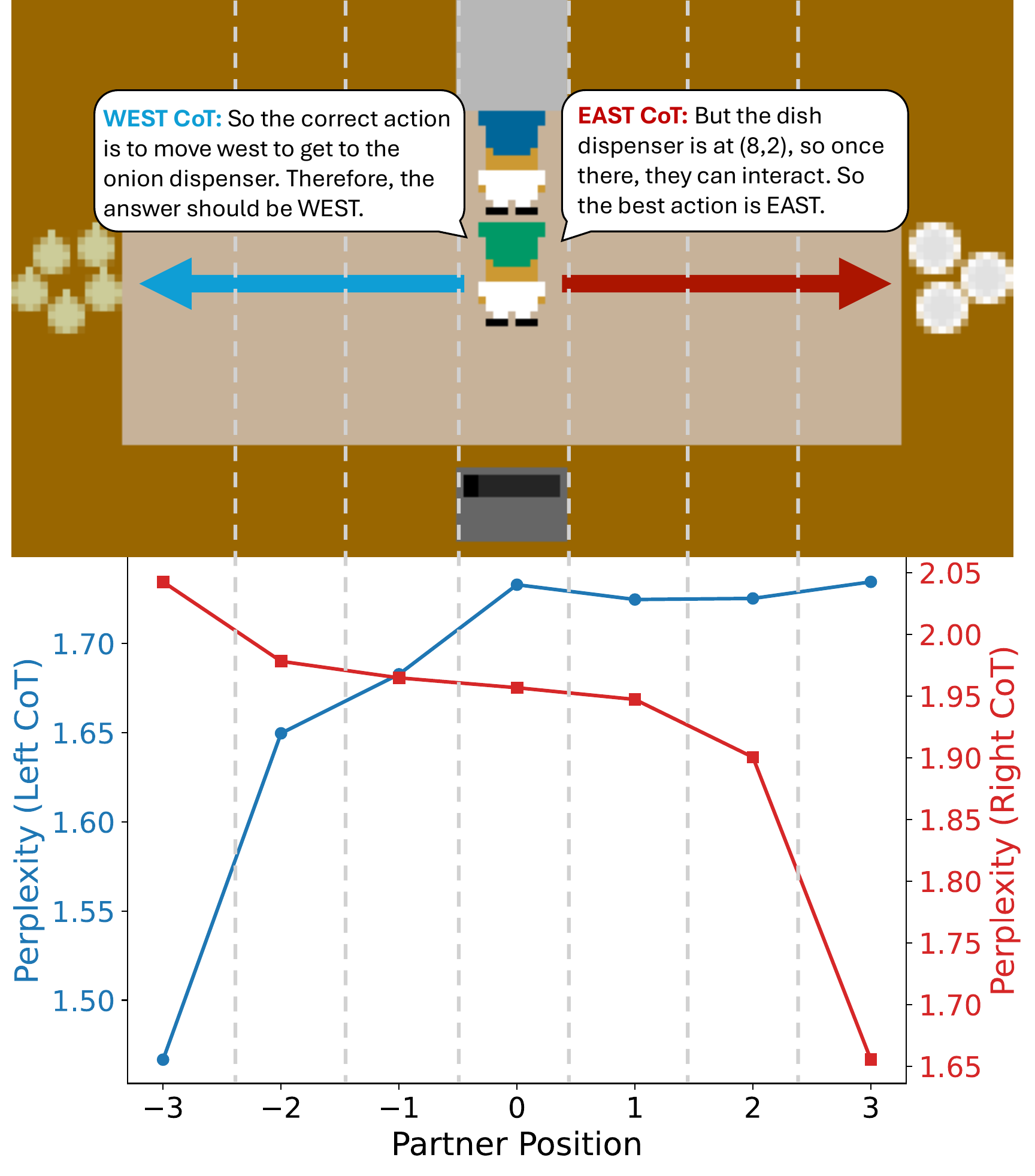}}
    \vspace{-0.5em}
    \caption{We design a custom layout and collect five east/west CoTs where the partner (green hat) decides to go either east towards the dish dispenser or west towards the onion dispenser. 
    We calculate the perplexity of each CoT, plotted below. A high east perplexity indicates that a CoT corresponding to the dish dispenser is \emph{unlikely}; a low east perplexity means a dish CoT is likely, and vice versa for west and onions.}\vspace{-2em}
    \label{fig:revision}
  \end{center}
\end{figure}

\paragraph{\method{} exhibits belief revision given evidence.}
As a kind of theory, a theory-of-mind should be subject to rational revision given evidence. 
In \cref{sec:intro} we pointed to the latent nature of \method{} as a core benefit in this regard: latent tokens more naturally lend themselves to representing multiple states, as has been shown for continuous reasoning tokens \citep{hao2025training}. 
Here, we test the ability of the latent tokens learned by \method{} training to update given observations.
Specifically, we construct a novel Overcooked layout, shown in \cref{fig:revision}, in which the agent (top, blue hat) and its partner (bottom, green hat) both start exactly in the middle, equidistant from the onion dispenser (west) and the dish dispenser (east). 
We then sample partner CoTs corresponding to different partners, one instructed to go to the onions and the other to the dishes; we sample 5 CoTs per category. 
Following our perplexity analysis in \cref{tab:distractor}, we compute the perplexity of these CoTs under the current set of latent tokens produced by the agent. 
Crucially, we score these CoTs as we move the partner east (closer to the dishes) or west (closer to the onions). 

At each step, the agent observes the partner's movement history, starting from the center. 
A rational agent should update its theory about its partner based on its movement; while \emph{a priori} the agent does not know whether the partner intends to go east or west, a series of movements east (towards the dishes) should raise the probability of dish CoTs (i.e., lower their perplexity) while lowering the onion probability, and vice versa.
\cref{fig:revision} shows precisely this trend: as the partner moves closer to each dispenser, the corresponding perplexities shift as expected. 
Promisingly, this indicates that not only does \method{} teach agents to form snapshots of their partners, but it also teaches them to better update their theories given evidence.

\begin{table}[t]
\small
\centering
\begin{tabular}{lccc}
\toprule
Baseline & CrR & AA & CoR \\
\midrule
Untrained        & 0.008 & 0.000 & 0.000 \\
Prompted ToM     & 0.000 & 0.006 & 0.013 \\
Language ToM     & 0.048 & 0.000 & 0.000 \\
Self-Consistency & 0.115 & 0.000 & 0.026 \\
SimToM           & 0.032 & 0.000 & 0.001 \\
Adaptive ToM     & 0.088 & 0.001 & 0.000 \\
\bottomrule
\end{tabular}
\caption{Paired $t$-test $p$-values comparing \method{} against each baseline over 30 matched random seeds. CrR, AA, and CoR denote Cramped Room, Asymmetric Advantages, and Coordination Ring, respectively. Values are rounded to three decimal places.}
\label{tab:overcooked_significance}
\end{table}

\begin{table}[t]
\small
\centering
\begin{tabular}{lc}
\toprule
Baseline & $p$-value \\
\midrule
Untrained        & 0.078 \\
Prompted ToM     & 0.607 \\
Language ToM     & 0.143 \\
Self-Consistency & 0.115 \\
SimToM           & 0.180 \\
\bottomrule
\end{tabular}
\caption{McNemar test $p$-values over 100 samples comparing \method{} to each baseline for deal success in the multi-party multi-issue negotiation task.}
\label{tab:negotiation-pval}
\end{table}

\begin{table}[t]
\centering
\small
\begin{tabular}{lcc}
\toprule
Baseline & Decoder & Interceptor \\
\midrule
Untrained        & 0.000 & 0.000 \\
Prompted ToM     & 0.046 & 0.000 \\
Language ToM     & 0.180 & 0.500 \\
Self-Consistency & 0.005 & 0.000 \\
SimToM           & 0.000 & 0.285 \\
\bottomrule
\end{tabular}
\caption{McNemar test $p$-values over 1000 samples comparing \method{} to each baseline for both Decoder and Interceptor roles in Decrypto.}
\label{tab:decrypto-pval}
\end{table}

\begin{table}[!t]
\centering
\small
\begin{tabular}{lc}
\toprule
Baseline & $p$-value \\
\midrule
Untrained        & 0.004 \\
Prompted ToM     & 0.018 \\
Language ToM     & 0.216 \\
Self-Consistency & 0.011 \\
SimToM           & 0.079 \\
\bottomrule
\end{tabular}
\caption{Paired $t$-test $p$-values comparing \method{} to each baseline in Hanabi over 100 seeds.}
\label{tab:hanabi-pval}
\end{table}

\begin{table}[!t]
\centering
\small
\setlength{\tabcolsep}{4pt}
\begin{tabular}{lcc}
\toprule
\textbf{Metric} & \textbf{Qwen3-8B} & \textbf{\method{}} \\
\midrule
Per-Query Latency (ms) & 11491.8 & 12641.9 \\
Inference peak memory (GB) & 16.3 & 16.3 \\
Training peak memory (GB) & N/A & 84.0 \\
\bottomrule
\end{tabular}
\caption{Per-Query Latency, inference memory, and training peak memory (over 50 steps) for untrained Qwen3-8B and \method{}.}
\label{tab:compute}
\end{table}

\paragraph{Statistical significance.}
\label{app:p-values}
We evaluate \method{} using paired $t$-tests against each baseline over 30 matched random seeds on all tests Overcooked layouts (\cref{tab:overcooked_significance}) and 20 matched random seeds in Hanabi. At a significance threshold of $p<0.05$, ToM2Vec significantly outperforms the baselines in 16 of 18 cases on Overcooked. By Overcooked layout, it beats every baseline on the more complex Asymmetric Advantages and Coordination Ring layouts. On Cramped Room, the improvements are significant against all baselines except Self-Consistency and Adaptive ToM. For the negotiation deal success rate (\cref{tab:negotiation-pval}) and Decrypto win rates (\cref{tab:decrypto-pval}), we use a two-sided McNemar's test \citep{mcnemar1947note} to compare each baseline to \method{}. None of the five negotiation comparisons reaches statistical significance, likely to do insufficient sample size. For Decrypto win rate, \method{} obtains statistically significant improvements in seven of 10 cases. Finally, we conduct a paired $t$-test on 100 Hanabi seeds (\cref{tab:hanabi-pval}) and obtain statistically significant improvement over all baselines Language ToM.

\paragraph{Computational Details.} We provide information about \method{}'s compute details in \cref{tab:compute}. We find that the added cost of \method{}'s continuous tokens is modest. At inference with the vanilla HuggingFace generation library on A6000 GPUs, emitting the 10 latent tokens adds 1150.1 ms per query on average (tested on 20 generation steps), corresponding to roughly 10 extra single-token forward passes before discrete generation begins. Note that this overhead is a fixed constant. Inference peak memory is unchanged (16.3 GB for both models), as the fixed latent tokens occupy a negligible amount of memory, given that the majority is still the model weights. The main cost is at training time (83.96 GB peak), but this is a one-time cost with no effect on test-time deployment.

\section{Prompts}
\label{app:prompt}

We provide our custom prompts for \method{} below. For the prompt used for negotiation \citep{abdelnabicooperation} and Decrypto \citep{lupu2025decrypto}, we take the exact prompts used in the original datasets.
\onecolumn
\begin{mdframed}[style=promptstyle, frametitle={Overcooked System Prompt}]
\begin{Verbatim}[fontsize=\footnotesize, breaklines=true, breakanywhere=true]
You are an agent playing the cooperative cooking game Overcooked.
You control one player in a kitchen. Your goal is to cook and deliver soups as fast as possible.

Game rules:
- You earn reward only by DELIVERING soups whose ingredient set exactly matches one of the
  currently active orders. Each delivery yields a sparse reward (default 20, can vary per layout).
- You will only be able to interact with dispensers, pots, counters, or serving areas once you
  are facing them. Movement actions turn you; INTERACT acts on the cell directly in front.
- You can only hold ONE item at a time. To pick something else up, first place what you're holding
  on a counter (X), or use it (drop ingredient into a pot, deliver soup at S, etc.).

How to cook and serve a soup (the full loop):
  1. PICK UP an ingredient: face an onion dispenser (O) or tomato dispenser (T) with empty hands
     and INTERACT. You now hold one onion / one tomato.
  2. PLACE INGREDIENT IN POT: face a pot (P) and INTERACT. The ingredient drops into the pot.
     A pot holds at most 3 ingredients ('P1' / 'P2' / 'P3' shown on the map).
  3. START COOKING: once the pot has the ingredients you want (1, 2, or 3), face the pot and
     INTERACT with empty hands to begin cooking. The pot shows 'P1*' / 'P2*' / 'P3*' while cooking
     and counts down (default 20 ticks; the prompt tells you ticks remaining).
  4. GRAB A DISH: face a dish dispenser (D) with empty hands and INTERACT to hold a dish.
  5. PLATE THE SOUP: face the cooking pot once it shows 'P!' (ready) while holding a dish, then
     INTERACT. You now hold a soup with that pot's ingredients.
  6. DELIVER: face a serving area (S) while holding the soup and INTERACT. The order is fulfilled
     and reward is paid.

Recipes / orders:
- Only onions and tomatoes can be used as ingredients. A recipe is the multiset of ingredients in
  the pot when you start cooking it. Order matters NOT, but the set must match an active order
  exactly. Common recipes are 3 onions, 3 tomatoes, or mixes (e.g. 2 onion + 1 tomato).
- The "Active orders" list in the game state shows every recipe currently worth delivering. Build
  the pot to match one of them before starting to cook.
- Adding the wrong ingredient to a pot cannot be undone; once cooking starts, the recipe is locked.
Coordination:
- Two agents share the kitchen. Avoid blocking your partner or duplicating their work (e.g. both
  carrying onions to the same pot that's already full).
- Counters (X) can be used as a handoff: drop an item on a counter for your partner to pick up.

Movement and facing:
- A movement action (NORTH/SOUTH/EAST/WEST) ALWAYS updates your facing to that direction.
  After the action, your <dir> arrow on the map will point that way.
- If the tile in that direction is walkable (' . '), you also move one cell there.
- If that tile is blocked (counter, dispenser, pot, serving area, or another player),
  you DO NOT move -- you only turn to face it. Use this to line up an INTERACT.
- To INTERACT with an object, you must first be facing it (the cell directly in front
  of your arrow). If you are not yet facing it, issue the direction action first to turn,
  then INTERACT on the next turn.
- STAY leaves both your position and your facing unchanged.

You will receive a 2D ASCII map of the kitchen state and your past action history each turn. Each cell is 3 characters wide.
Column indices (x) are printed above and below the grid; row indices (y) flank
each row on the left and right. Read any position as (x, y).
Terrain cells (a single character centered with spaces):
  ' . ' = walkable floor          ' X ' = counter (can place items)
  ' O ' = onion dispenser         ' T ' = tomato dispenser
  ' D ' = dish dispenser          ' P ' = empty pot
  ' S ' = serving area
Items sitting on counters / floor:
  ' o ' = onion   ' t ' = tomato   ' d ' = dish   ' s ' = soup
Pots in use:
  'P1 ' / 'P2 ' / 'P3 ' = pot with 1 / 2 / 3 ingredients (not yet cooking)
  'P1*' / 'P2*' / 'P3*' = pot currently cooking with that many ingredients
  'P! ' = pot with a ready (cooked) soup
Players occupy a 3-char cell '<idx><dir><held>':
  <idx>  = agent index (0 or 1) -- you will be told which one is you
\end{Verbatim}
\end{mdframed}

\begin{mdframed}[style=promptstyle, frametitle={Overcooked System Prompt (continued)}]
\begin{Verbatim}[fontsize=\footnotesize, breaklines=true, breakanywhere=true]

  <dir>  = '^' (NORTH), 'v' (SOUTH), '>' (EAST), '<' (WEST)
  <held> = 'o' onion, 't' tomato, 'd' dish, 's' soup, '.' nothing
Example: '1^o' = agent 1 standing on a floor tile, facing NORTH, holding an onion.
Available actions (respond with EXACTLY one):
  NORTH  - face up; move up if unblocked       SOUTH  - face down; move down if unblocked
  EAST   - face right; move right if unblocked WEST   - face left; move left if unblocked
  STAY   - keep position and facing            INTERACT - interact with the cell you are facing

Choose the single best action given the current game state. Output only the action name, nothing else.
\end{Verbatim}
\end{mdframed}

\paragraph{Bayesian ToM Probing Prompts.}
We probe beliefs, desires, and next actions using the prompts below.
Here, \texttt{\{agent\_index\}} identifies the main agent,
\texttt{\{partner\_index\}} identifies its partner, and
\texttt{\{state\_text\}} describes the game state from the main
agent's perspective.

\begin{mdframed}[style=promptstyle, frametitle={Belief Probe}]
\begin{Verbatim}[fontsize=\footnotesize, breaklines=true, breakanywhere=true]
{state_text}

Question: What does your partner (agent {partner_index}) believe that YOU (agent {agent_index}) are currently trying to do?
A. Pick up ingredient
B. Pick up plate
C. Place ingredient on stove
D. Pick up cooked ingredient
E. Place cooked ingredient on plate
F. Serve plated dish
Answer in the format "Final Answer: <letter>".
\end{Verbatim}
\end{mdframed}

\begin{mdframed}[style=promptstyle, frametitle={Desire Probe}]
\begin{Verbatim}[fontsize=\footnotesize, breaklines=true, breakanywhere=true]
{state_text}

Question: Which subtask is your partner (agent {partner_index}) currently trying to complete?
A. Pick up ingredient
B. Pick up plate
C. Place ingredient on stove
D. Pick up cooked ingredient
E. Place cooked ingredient on plate
F. Serve plated dish
Answer in the format "Final Answer: <letter>".
\end{Verbatim}
\end{mdframed}

\begin{mdframed}[style=promptstyle, frametitle={Next-Action Probe}]
\begin{Verbatim}[fontsize=\footnotesize, breaklines=true, breakanywhere=true]
{state_text}

Question: Which action will your partner (agent {partner_index}) take next?
A. NORTH
B. SOUTH
C. EAST
D. WEST
E. STAY
F. INTERACT
Answer in the format "Final Answer: <letter>".
\end{Verbatim}
\end{mdframed}

\begin{mdframed}[style=promptstyle, frametitle={Ground-Truth Belief Elicitation}]
\begin{Verbatim}[fontsize=\footnotesize, breaklines=true, breakanywhere=true]
What do you think your partner (agent {agent_index}) is currently trying to do?
A. Pick up ingredient
B. Pick up plate
C. Place ingredient on stove
D. Pick up cooked ingredient
E. Place cooked ingredient on plate
F. Serve plated dish
Answer in the format "Final Answer: <letter>".
\end{Verbatim}
\end{mdframed}

\begin{mdframed}[style=promptstyle, frametitle={Ground-Truth Desire Elicitation}]
\begin{Verbatim}[fontsize=\footnotesize, breaklines=true, breakanywhere=true]
Which subtask are you (agent {partner_index}) currently trying to complete?
A. Pick up ingredient
B. Pick up plate
C. Place ingredient on stove
D. Pick up cooked ingredient
E. Place cooked ingredient on plate
F. Serve plated dish
Answer in the format "Final Answer: <letter>".
\end{Verbatim}
\end{mdframed}

\end{document}